\documentclass[letterpaper, 10 pt, conference]{ieeeconf}
\IEEEoverridecommandlockouts

\usepackage[utf8]{inputenc}
\usepackage{cite}
\usepackage{amsmath,amssymb,amsfonts}
\usepackage{algorithmic}
\usepackage{graphicx}
\usepackage{subfigure}
\usepackage{gensymb}
\usepackage{textcomp}
\usepackage{float}
\usepackage{dblfloatfix}    
\usepackage{xcolor}
\usepackage{comment}
\usepackage{geometry}
\usepackage{longtable}
\usepackage{tabularx,booktabs}

\usepackage[export]{adjustbox}
\usepackage{wrapfig}

\title{Contact-less Manipulation of Millimeter-scale Objects via Ultrasonic Levitation}

\author{\authorblockN{Jared Nakahara$^{1}$, Boling Yang$^{2}$, and Joshua R. Smith$^{1,2}$}
\authorblockA{$^{1}$Dept. of Electrical and Computer Engineering and $^{2}$Paul G. Allen School of Computer Science and Engineering}}

\date{September 2019}

\begin{document}

\maketitle

\begin{abstract}
Although general purpose robotic manipulators are becoming more capable at manipulating various objects, their ability to manipulate millimeter-scale objects are usually very limited. On the other hand, ultrasonic levitation devices have been shown to levitate a large range of small objects, from polystyrene balls to living organisms. By controlling the acoustic force fields, ultrasonic levitation devices can compensate for robot manipulator positioning uncertainty and control the grasping force exerted on the target object. The material agnostic nature of acoustic levitation devices and their ability to dexterously manipulate millimeter-scale objects make them appealing as a grasping mode for general purpose robots. In this work, we present an ultrasonic, contact-less manipulation device that can be attached to or picked up by any general purpose robotic arm, enabling millimeter-scale manipulation with little to no modification to the robot itself. This device is capable of performing the very first phase-controlled picking action on acoustically reflective surfaces. With the manipulator placed around the target object, the manipulator can grasp objects smaller in size than the robot's positioning uncertainty, trap the object to resist air currents during robot movement, and dexterously hold a small and fragile object, like a flower bud. Due to the contact-less nature of the ultrasound-based gripper, a camera positioned to look into the cylinder can inspect the object without occlusion, facilitating accurate visual feature extraction.
\end{abstract}

\section{Introduction}
\label{Sec:Introduction}

\begin{figure}[!t]
    \centering
    \setlength{\abovecaptionskip}{0pt}
    \includegraphics[width=0.48\textwidth]{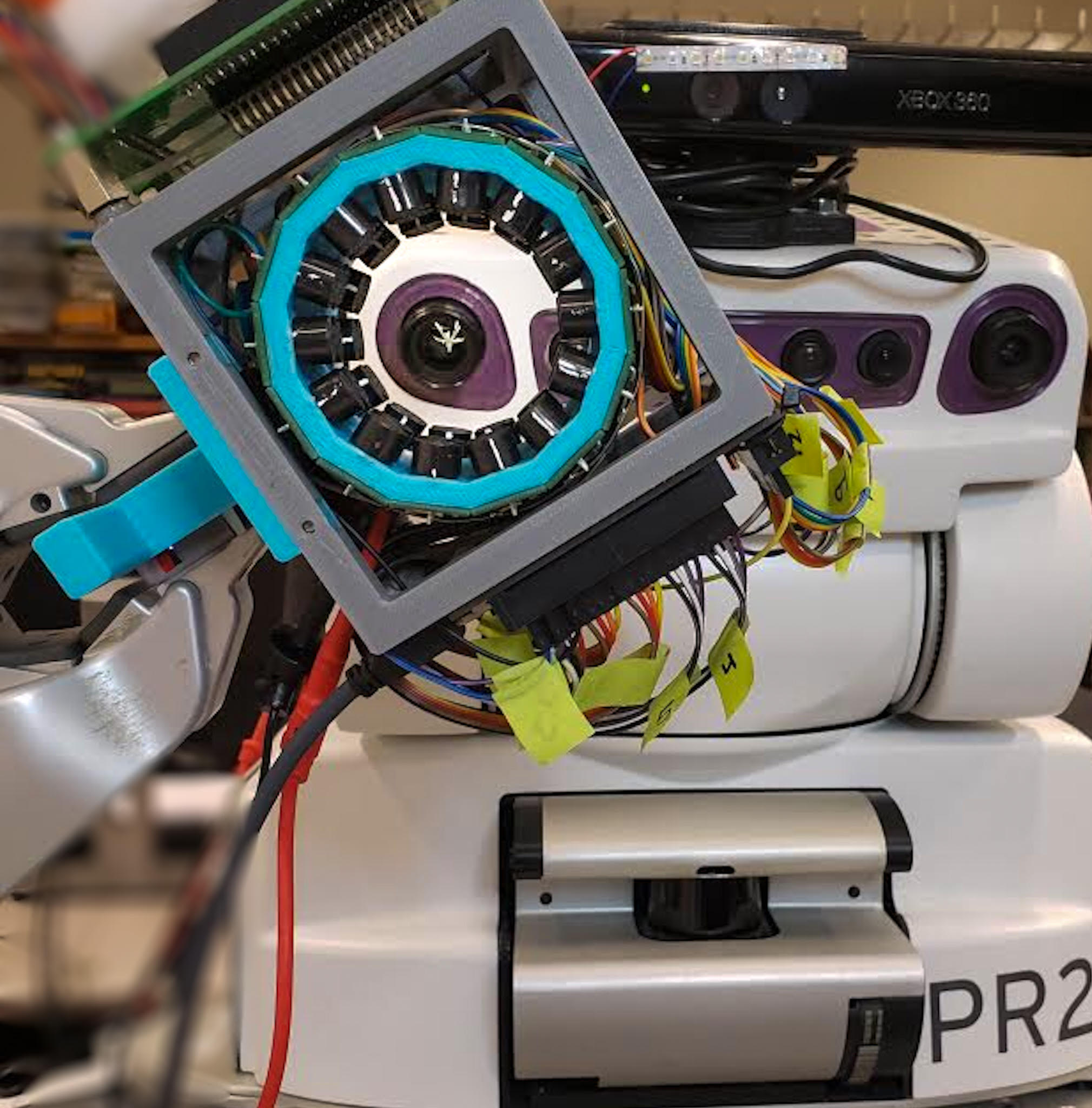}
    \caption{The PR2 robot is using the contact-less ultrasonic manipulator to hold a flower.}
    \label{DoubleRing}
\end{figure}

Over the past decade, robots have become increasingly capable of dexterous manipulation, accurate object pose estimation, sophisticated motion planning, and improved motor skills\cite{ma2011dexterity}, typically achieved by arm and manipulator designs with high degrees of freedom. However, in most general purpose robots, dexterity is limited to manipulating larger objects on the scale of centimeters or above, lacking the ability to perform the same level of manipulation on smaller objects. With limited gripping force resolution and positioning accuracy, these robots can miss or damage the target object in cases when the object of interest is small or fragile. However, millimeter-scale manipulation is a skill required for daily life, scientific research, and in the manufacturing industry. For example, in biology, neural science, and other related research areas, experiments that involve handling small, deformable and fragile objects like insects\cite{bolling2015insect,batchelor2019sound,gu2007whole}, biological tissues\cite{levy1996retroviral, stefanoff2003laboratory}, and drops of fluid\cite{kong2012automatic,song2006reactions}, are very common. In microassembly and PCB manufacturing industries, objects like bare silicon dyes and electronic components are assembled with machines that require an extra level of precision because of their size and fragility. The increasing need to automate and improve these experiments and manufacturing processes \cite{cork2006advanced,miles2018achieving,melin2007microfluidic,chollet1999cost,bohringer1998parallel,heriban2008robotic} highly motivates further research into enabling general purpose robots to manipulate millimeter-scale objects.   

Acoustic levitators and traps have demonstrated the capability to manipulate objects on the scale of millimeters through the processes of localized pressure modulation via high frequency acoustic wave interference, at frequencies inaudible to the human ear. By taking advantage of the non-linearity of sound in air, sound can be used to generate lift on an object within the acoustic field, allowing it to overcome gravitational forces. In this work, we present an acoustic levitation device that allows general purpose robots to pick and manipulate millimeter-scale objects. This paper makes the following contributions:
\begin{itemize}
    \item Development of the first acoustic levitation device and control scheme that can pick up an object from a table top
    \item Implementation of an acoustic manipulator that can be attached to or be picked up by general purpose robots with little to no robot modification
\end{itemize}
While other publications have shown that small objects can be acoustically levitated and manipulated, the contact-less ultrasonic robotic manipulator is the first device to enable a general purpose robot to pick up small millimeter-scale objects from an acoustically reflective surface, such as a table top.

This acoustic manipulator's size and weight allow it to be easily attached to most general purpose robots. The levitation device can manipulate objects that are placed within an area called the \textit{basin of attraction}. This large manipulation area adds extra robustness to compensate for robot positioning uncertainty.

In addition, we use an acoustic field modeling method, called the method of images, to improve the simulated model of force dynamics inside the levitator. This modeling method improves the dexterity of our acoustic manipulator by enabling novel acoustic manipulation skills including picking object from a flat surface without the need for object support structures or specialized dispensing of the object. 

The manipulator was tested for real world usability via a PR2 robot, where the PR2 attempted to pickup an object, extract visual features from the object, and perform object sorting based on the extracted features. We found that the acoustic levitator allows general purpose robots to manipulate millimeter scale objects, by providing ``in-hand" dexterity and enhanced tolerance to the robot's kinematic uncertainty. Further more, the contact-less nature of this manipulator provides an unblocked view of the whole object. To the best of our knowledge, this is the first research that has successfully enabled a PR2 robot (or robots of similar form factors) to pick up and inspect small and fragile objects like mosquitoes, integrated circuit chips, and flowers without damaging them.


\section{Related Work}
\label{Sec:RelatedWork}
In this section, we will review the previous work on enabling high accuracy and gentle manipulation from the perspective of general purpose, industrial, and medical robotics, and discuss previous work in acoustic levitation devices.

To manipulate small objects, a robot should have sufficient manipulation accuracy and dexterity. Although dexterous manipulation has been a popular research topic, exploration of millimeter scale object manipulation on general purpose robotic platforms has been extremely limited. Many groups have developed solutions to reduce the positioning uncertainty and increase dexterity of general purpose robots,\cite{yang2017pre,johannsmeier2019framework,odhner2014compliant}, yet none of the objects tested have volumes close to millimeter scale. Industrial robotic arms are known for having great positioning accuracy, equipped with specialized manipulators and sensors, and are capable of the automatic assembly of small screws \cite{aronson2016data,jia2018survey,cheng2018sensor}. However, object fragility has not been considered by any of the research above.

In contrast, millimeter scale manipulation is very common in the field of surgical robotics \cite{le2016survey}. Surgical manipulators have great manipulation accuracy and dexterity\cite{kutzer2011design,murphy2014effects,aoyama2014precise}, yet their actuation system, kinematics and end-effectors are designed for surgical scenarios. This level of specialization could make transferring design concepts to general purpose robots difficult due to the broadness of tasks these robots are required to perform.


On the other hand, robots can handle fragile objects through constant object state estimation and adjusting grasping strategy accordingly. One effective way to monitor object state through a manipulation process is to mount sensors on the end-effectors of a robot. Lancaster et. al. \cite{lancaster2019improved} created a fingertip-mounted sensor that provides proximity, contact detection, and force sensing. Koyama et. al. \cite{koyama2018high} designed a proximity sensor that allows a robot hand to catch a falling paper balloon with insignificant deformation. Lee et. al. \cite{lee2019soft} proposed a soft linear actuator allowing a robot hand to hold a potato chip without crushing it. Likewise, development of soft robotic grippers is another popular research direction toward gentle manipulation \cite{gupta2016learning,rus2015design,galloway2016soft,ilievski2011soft,hughes2016soft}. However, the above methods have only been tested on larger objects and will fail to manipulate millimeter scale objects. 

Acoustic levitation devices use a propagating pressure wave to exert a lifting or trapping force on an object.  Xie \cite{xie2006acoustic} showed that a single-axis acoustic levitator was capable of lifting bio-materials, small living organisms, and tissue, in air, without damage. Some traps are capable of performing dexterous manipulation of objects in air \cite{seah2014correspondence} and water \cite{courtney2013dexterous} using Bessel functions. Others are capable of pseudo-picking actions \cite{marzo2016gauntlev}, translation, and rotation \cite{marzo2015holographic} of objects using coordinated phase delays. Kozuka et. al. \cite{kozuka2007noncontact} showed that small objects could be manipulated in air by controlling the phase of two interfering sound beams. In these devices, objects were placed inside the levitator and objects were translated using a combination of phase delays and acoustic source axes deflections. It was also found that slight variations in acoustic source frequencies could cause object translation.

Other applications of acoustic levitation based manipulators include display technology, \cite{ochiai2014three}, and a microassembly device, \cite{youssefi2019contactless}. These devices use large arrays of phase controlled transducers to translate small particles, thereby demonstrating that acoustic levitation could be used for complex three-dimensional manipulation of levitated small particles, such as polystyrene balls. While these systems perform object translation and rotational control, both the transducer arrays and device are large and would not be feasible as a robotic manipulator attachment. 

Using a similar multi-transducer array, Marzo \cite{marzo2016gauntlev} presents a hand mountable device (GauntLev), that uses a planar transducer array and phase manipulation to translate and rotate objects. Additionally, GauntLev is capable of performing a picking action by first trapping a particle within a standing wave and then lifting the particle by hand to a higher vertical position. The device and particle can then be moved away from the acoustic reflector. Due to the one-sided array geometry, the levitating force is low when compared to a double-sided array geometry, limiting the types of objects that can be manipulated. The device also requires the object to be move by hand to achieve object picking which is not desirable for a robot mounted manipulator.
\section{Principles of Operation}
\label{Sec:Principles of Operation}

\begin{figure}[b!]
    \centering
    \setlength{\abovecaptionskip}{0pt}
    \includegraphics[width=0.40\textwidth]{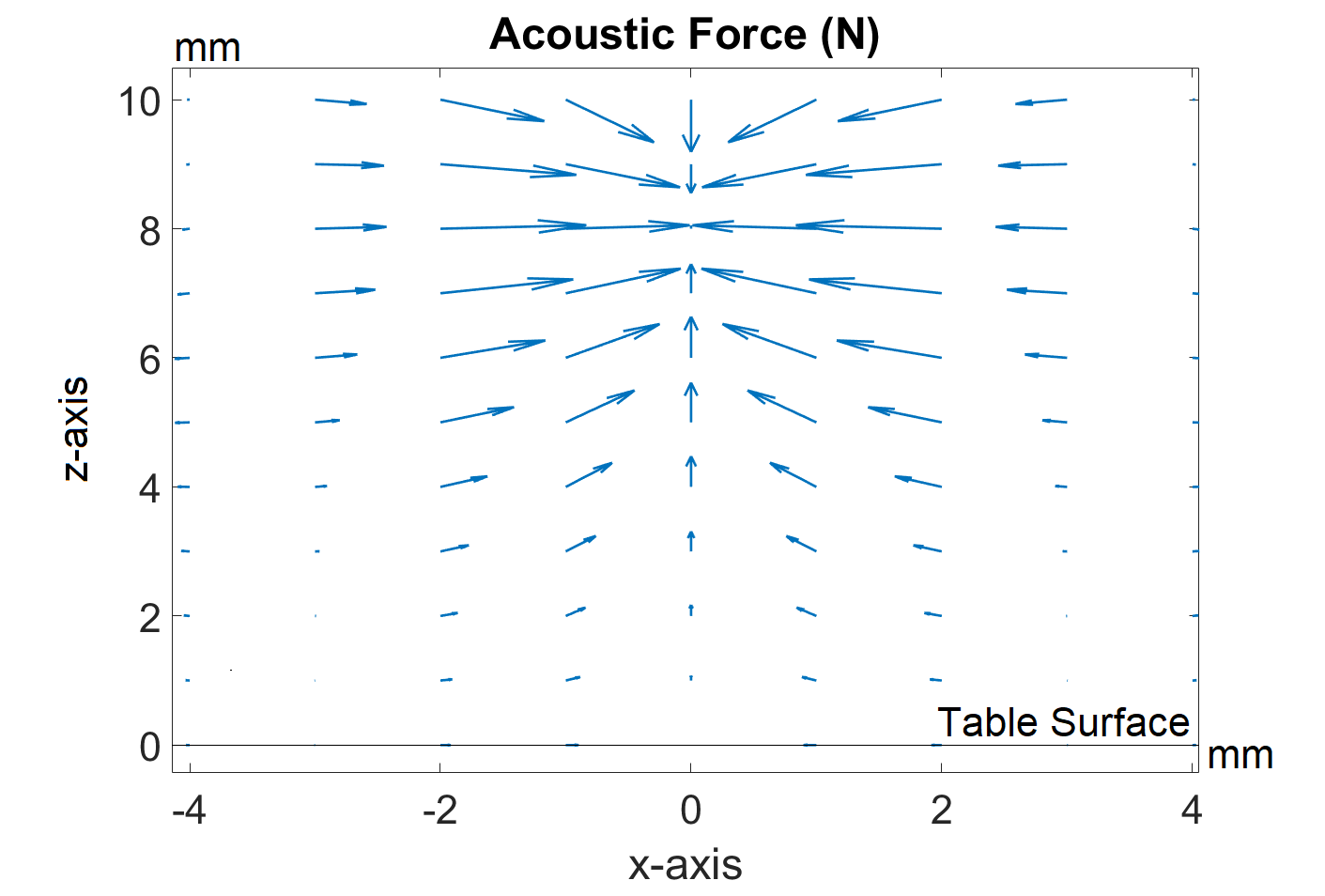}
    \caption{Quiver plot showing the simulated acoustic force field exerted on an object 2mm in diameter during object picking. The object is pulled upward to the stable convergence point located 8mm above the table surface. }
    \label{pickingQuiver}
\end{figure}

\subsection{Acoustic Levitation and Trapping Force}

Acoustic levitation devices manipulate objects by modulating air particles using ultrasonic waves. The ultrasonic waves create a spatially distributed, time average acoustic energy pattern. Objects inside the acoustic field will move from positions of high acoustic potential energy, to areas of low acoustic potential energy. Based on the transducers used and the voltage applied to the transducers, the acoustic pressure and air velocity distribution can be calculated. Given the acoustic pressure distribution ($p$), and air velocity distribution ($v$), the acoustic potential energy or Gor'kov potential ($U$) can be found by applying the expression in Eq.\eqref{gorkov}, \cite{bruus2012acoustofluidics},

\begin{equation}
    U = 2\pi R^3 \left( \frac{ \left< p^2 \right>}{3 \rho_o c_o^2} - \frac{\rho_o\left< v \cdot v \right>}{2} \right)
    \label{gorkov}
\end{equation}
where $R$ is the radius of the levitated particle, $\rho_o$ is the density of air and $c_o$ is the speed of sound in air.

\begin{figure}[t!]
    \centering
    \setlength{\abovecaptionskip}{0pt}
    \includegraphics[width=0.46\textwidth]{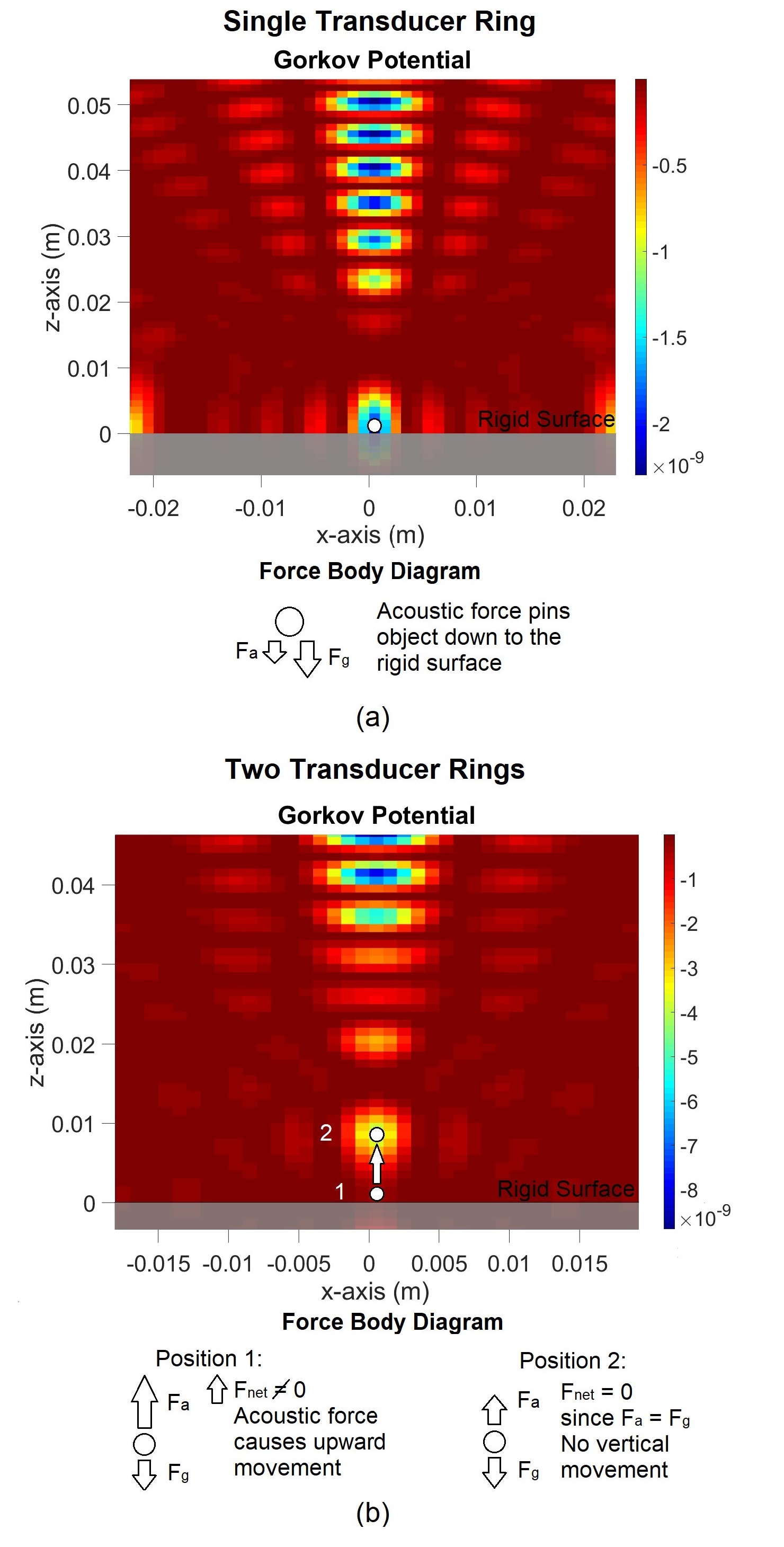}
    \caption{(a) A Gor'kov potential distribution incapable of lifting an object due to the stable node located on the rigid surface. A particle initially resting on the table top experiences a downward acoustic force, pinning it to the table. This potential is due to a single ring of transducers above a rigid reflecting surface. (b) A Gor'kov potential distribution capable of lifting an object off a table top. Under the influence of this potential, a particle that is initially resting on the table top (position 1) experiences a net upward force. As illustrated in the force-body diagram, at position 1 the upward force due to the acoustic field exceeds the downward force of gravity. Once this potential is enabled by turning on the ultrasound field, the particle rises until it reaches position 2, a stable local minimum at which gravity balances the upward acoustic force. This potential is due to two stacked rings of transducers above a rigid reflecting surface. }.
    \label{GorkovForceBody}
    \vskip -6mm
\end{figure}


The acoustic force is the gradient of the Gor'kov potential. From the Gor'kov acoustic potential scalar field, the acoustic force vector field ($F_a$) can be calculated using Eq. \eqref{acoustic_force}. 
\begin{equation}
    F_a = -\nabla U
    \label{acoustic_force}
\end{equation}
The force vector convergence, Fig.\ref{pickingQuiver}, and force body diagrams, Fig.\ref{GorkovForceBody}, show the acoustic force dynamics experienced by the trapped object. The movement of the object from position 1 to 2 is enabled by the levitator geometry which will be discussed in a later section. The resting position of the object is located in mid air at a local minima of the Gor'kov potential where gravity ($F_g$) equals the upward acoustic lift force.

\subsection{Modeling Acoustic Field}

Since the picking task requires an object to be lifted off an acoustically reflective surface, both the incident and reflected acoustic waves must be modeled to accurately represent the acoustic field and predict object behavior. 
The individual transducers are modeled as sources that superimpose linearly. The rigid table surface is a boundary at which the particle velocity normal to the table and acoustic pressure of the air must be zero, since the air cannot penetrate the table.  

The rigid table surface is analogous to a ground plane in electrostatics, located at the x-y plane in Fig. \ref{pickingQuiver} and \ref{GorkovForceBody}. In electrostatics, point sources in the vicinity of a ground plane can be modeled very effectively via the method of images. In the method of images, the ground plane is removed, but its effect is modeled exactly by a new set of so-called ``image'' charges. If a real charge of amplitude $q$ is at coordinates $(x,y,z)$, its image has charge $-q$ and is placed at $(x,y,-z)$. By symmetry the $z$ components of the field cancel, and thus the image charges emulate the effect of the ground plane.

The method of images can also be applied to acoustics, as in for example \cite{Gaunaurd1994acoustic}. In our acoustic case, we introduce a set of image transducers, mirrored below the plane. Because of symmetry considerations similar to the electrostatic case, the particle velocity contributions of each real and mirror transducer cancel in the normal direction at the table surface, satisfying the boundary condition of a rigid surface. 

The method of images allows us to efficiently compute the Gor'kov potential and resulting force field while considering the boundary condition for the transducers in close proximity to the reflective surface. Simply neglecting the reflective surface provides qualitatively incorrect results. Correctly modeling and controlling the reflections from the table surface is essential for being able to successfully lift objects off of the surface.

\subsection{Lifting Objects from Rigid Surfaces}

From the models of the Gor'kov potential, the acoustic field must be shaped such that an upward vertical lift force is generated to lift the object off of the rigid surface. One initial approach would be to create a single ring of inward facing transducers of the same height to generate a stable Gor'kov potential node in the center of the ring just above the target object. However, due to the reflection of the acoustic field off the rigid surface, a stable Gor'kov node would form directly on the target object, pinning the object down on the rigid surface, rather than lifting the object up, Fig. \ref{GorkovForceBody}a. This is due to the reflected acoustic field originating from the imaginary sources located below the acoustically reflective boundary. As a result, the Gor'kov node is placed on the reflective boundary at a height midway between the real and imaginary acoustic source heights, at a height of zero. 

To place a Gor'kov node above the target object, like in Fig. \ref{GorkovForceBody}b, at least two rings of inward facing transducers stacked on top of each other are needed. Like in the single ring of transducer example, Gor'kov nodes are created between the respective heights of each transducer ring, including the imaginary rings. The second transducer ring cancels the Gor'kov node located on the reflective surface. This creates a region of vertical instability at the reflective boundary, pushing the object into a stable Gor'kov node as depicted in Fig. \ref{GorkovForceBody}b.
\begin{figure}[t!]
    \centering
    \setlength{\abovecaptionskip}{0pt}
    \includegraphics[width=0.48\textwidth]{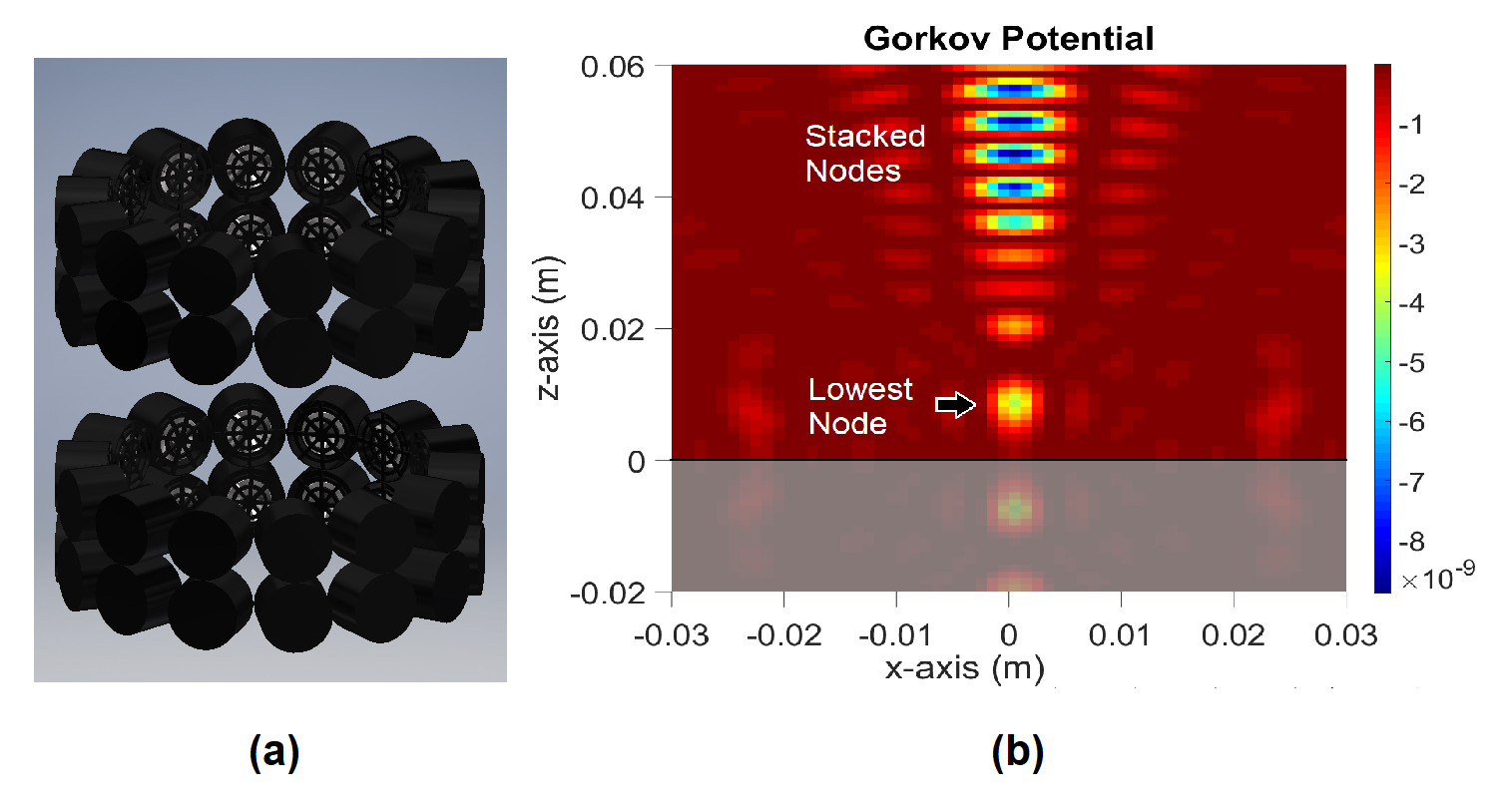}
    \caption{(a) The ultrasonic transducers are arranged in a cylinder consisting of four ring levels, numbered 1 to 4 from bottom to top. (b) The four level ring geometry generates a stacked node structure with stable points located along the axis of the cylinder. These stable points trap objects and can be easily translated along the central axis of the cylinder. The acoustically reflective boundary is shown as the black horizontal line.}
    \label{ringAndNode}
    \vskip -6mm
\end{figure}

\begin{figure}[b!]
    \centering
    \setlength{\abovecaptionskip}{0pt}
    \includegraphics[width=0.40\textwidth]{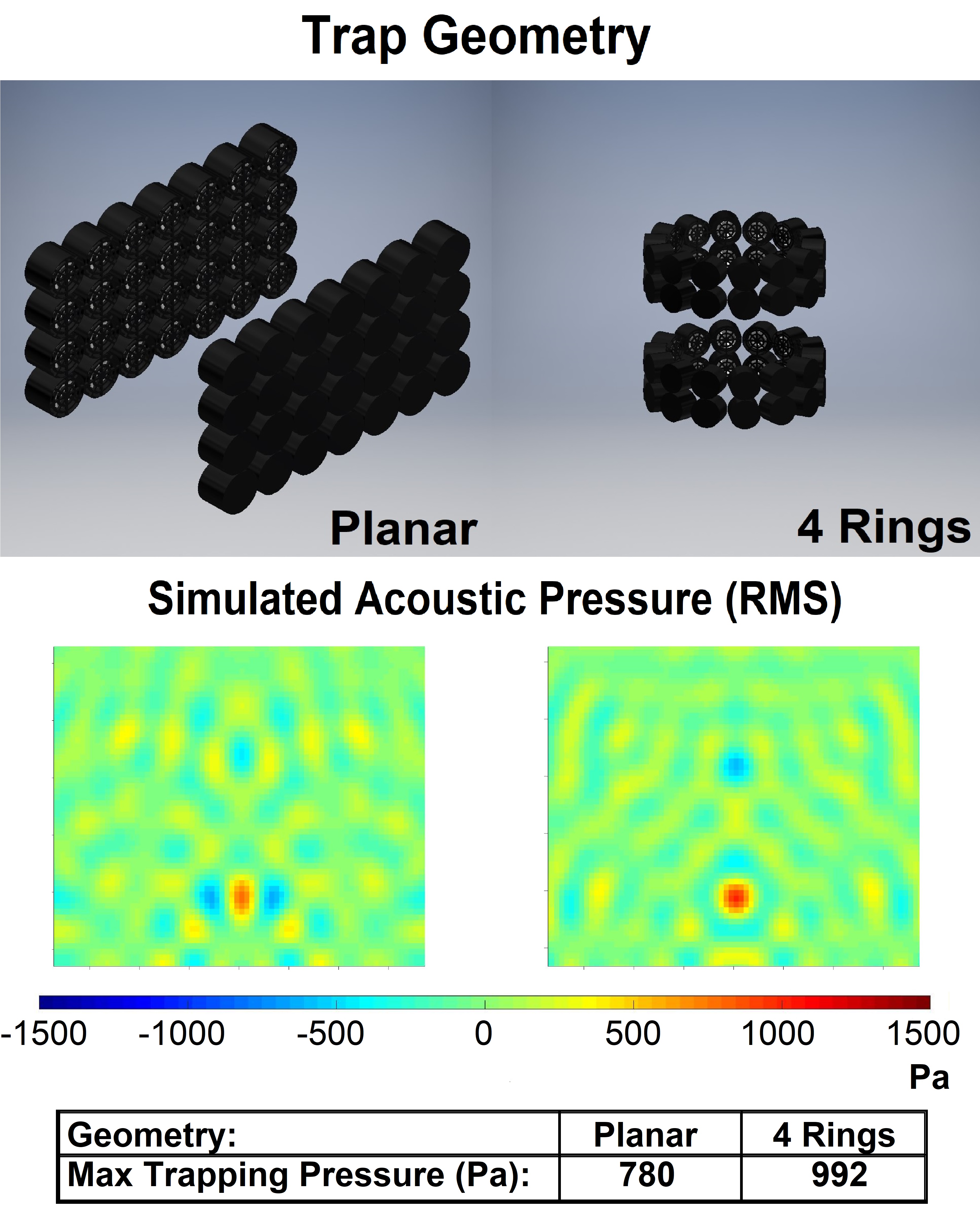}
    \caption{A comparison between a planar and cylindrical trap geometry with respect to the maximum acoustic pressure delivered to a focused target location. The cylindrical geometry is capable of delivering 27\% more pressure than the planar geometry.}
    \label{geometryComp}
\end{figure}

 \begin{figure*}[t!]
    \centerline{\includegraphics[width=\linewidth]{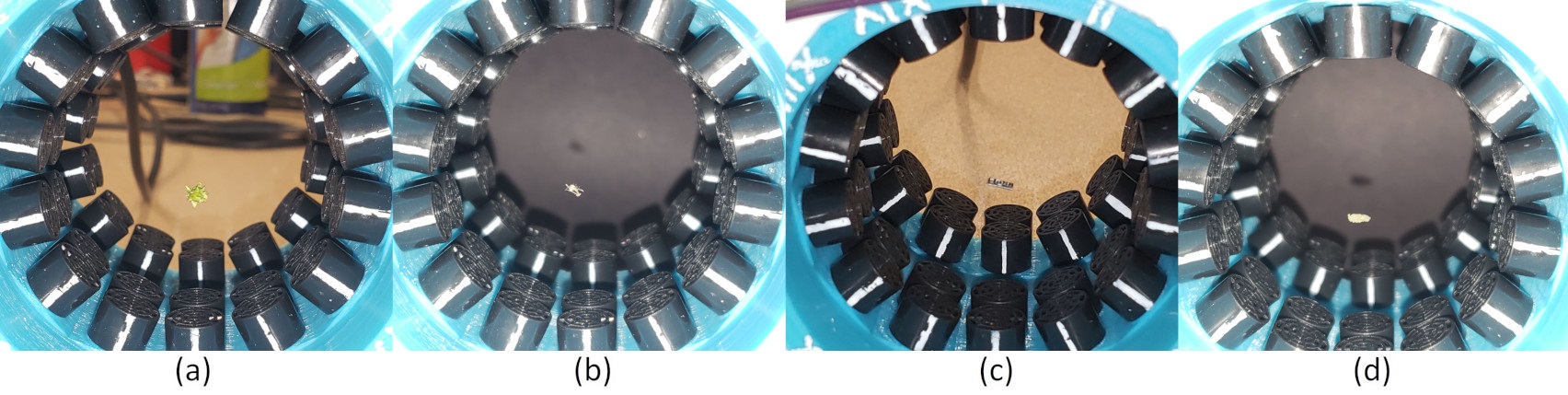}}
    \setlength{\abovecaptionskip}{0pt}
    \caption{The ultrasonic levitation device is a material agnostic manipulator capable of suspending (a) plant pieces, (b) small insects [mosquito], (c) integrated circuits, and (d) food [potato chip crumb]. Unlike magnet based levitators, the levitated object is not restricted to specific materials. Objects stay suspended in the middle of the trap, with the force of gravity directed downward in the frame of the pictures.}
    \label{ItemsLevitated}
    \vskip -2mm
\end{figure*}

\begin{table*}[hb!]
    \centering
    \vspace{-4mm}
    \caption{Mass of objects picked up by device}
    \vspace{-2mm}
    \begin{tabular}{|c||c|c|c|c|c|c|}
    \hline
    Object & Flower Bud &  Plant Piece & Mosquito & IC & Potato Chip & Red Polystyrene Ball \\
    \hline
    Weight (mg) & 3 & 3 & 3 & 12 & 5 & 1\\
    \hline
    \end{tabular}
    \label{objectMassTable}
\end{table*}

\section{System Design}
\label{Sec:SystemDesign}

 The ultrasonic levitation manipulator is specifically designed to pick small, fragile objects up off of acoustically reflective surfaces through transducer phase manipulation. This allows robots such as the PR2 to gain additional dexterity through contact-less object manipulation. To maximize trap stability and acoustic force, the trap leverages both careful selection of the array geometry and phase manipulation of the transducer array. By using a cylindrical geometry rather than a planar geometry, acoustic energy is focused along the cylindrical axis and concentrated into a smaller trapping volume rather than being dispersed.

\subsection{Acoustic Trap Geometry}

To help maximize trapping force and enable dexterous manipulation, an inward facing cylindrical array of 56 transducers arranged into four rings, Fig. \ref{ringAndNode}a, aims to exert a higher degree of force and lift objects further away from the surface of the table than a planar transducer design. To perform the picking action, more than one ring is need to correctly shift the Gor'kov potential. While only the bottom two rings of the manipulator are necessary for lifting an object off of a reflective surface, the top two rings lift the object further off of its initial resting surface and secure the object at a higher position. This reduces the effect of the imaginary acoustic sources on the object, allowing the whole device to be lifted without affecting the object.

A cylindrical geometry allows all transducers to be positioned equidistantly from a central axis with pairs of transducers facing each other. The transducers form stacked rings of a diameter five times the acoustic wavelength, 8.5mm. When activated, the trap generates multiple 40kHz standing waves which intersect to form a stacked node structure, Fig. \ref{ringAndNode}b, along the central axis of the cylinder. This concentrates the acoustic energy along the central axis of the device and aids in the device's specialized task of vertical object picking and lifting allowing an object to be translate upward from node to node. In order to get a better estimate of relative grasping pressure, Fig. \ref{geometryComp} compares the picking node pressure between a 56 transducer planar array and a cylindrical array.

\subsection{Hardware Design}

To generate the acoustic field, a field programable gate array, FPGA, generates 56 channels of individually phase controlled square wave outputs. Each output channel has a phase resolution of 2500 steps or approximately 0.15$^{\circ}$. The FPGA allows all output channels to be updated on the same clock edge for precise transducer clock timings. The FPGA outputs are level translated from 3.3V logic to 12V logic and a second signal with inverted phase is generated. The original and phase inverted signals drive the 56 MA40S4S transducers differentially such that the transducers see a $24V_{pp}$ signal. 

\subsection{Trap Control System}

To control the ultrasonic manipulation device, the acoustic field's low potential nodes must be spatially translated. The field is changed by manipulating the phase of each transducer output channel through either a increment, decrement, or phase angle set command sent from a computer to the FPGA. The commands are decoded by the FPGA and the phase values for the next state of each channel are buffered. Once the buffer is full with the next set of phases, a synchronous update occurs to update the output channels at the same time. This system allows the ultrasonic manipulation device to be manually controlled by a human, through increment and decrement commands, or by a computer program that can compute and send all 56 phase angles for a given position.

\subsection{Simulating the Design}

To validate the picking action of the acoustic field, a MATLAB simulation renders the acoustic field corresponding to the activation of the first and second transducer ring levels. Fig. \ref{pickingQuiver} shows the vertical acoustic force as an object is picked from the table surface to a stable levitation node about 8mm above the table surface. If the object is subjected to a lateral perturbation, left or right, the object will return to the stable levitation node due to the lateral trapping force depicted by the arrow convergence.

\begin{figure*}[ht!]
    \centering
    \setlength{\abovecaptionskip}{0pt}
    \includegraphics[width=\linewidth]{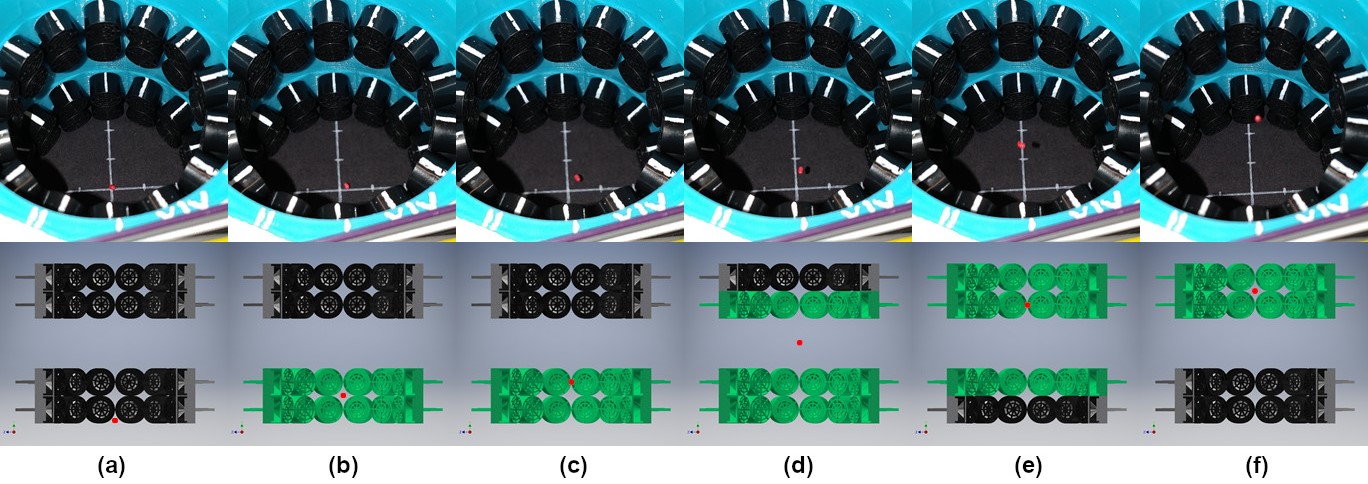}
    \caption{The picking action is performed in 6 steps. In the figure, green represents activated transducers involved in the picking process. (a) The manipulator is placed around the target object and brought to the center of the basin of attraction. (b) The first two levels of the transducer rings are activated and an incremented phase shift is applied to the second level of transducers, bringing the object to about 10mm off the acoustically reflective surface of the table. (c) An incremental phase shift is applied to the bottom transducer array, moving the object from 10mm to 15mm. (d) The third level ring of transducers are activated and the phase of the second and third level transducers are incremented. This moves the object from 15mm to 30mm in height. (e) The first level transducers are deactivated, and the fourth level transducers are activated. The phase of the third and fourth, transducer ring levels are incremented, moving the object from 30mm in height to 45mm in height. (f) The second level transducer ring is deactivated. The fourth level transducer ring phase is decremented. At this height, 50mm, the levitation device can be lifted from the surface of the table.}
    \label{MultiFramePicking} 
    \vskip -6mm
\end{figure*}

\begin{figure}[b!]
    \centering
    \vspace{-4mm}
    \setlength{\abovecaptionskip}{0pt}
    \includegraphics[width=0.35\textwidth]{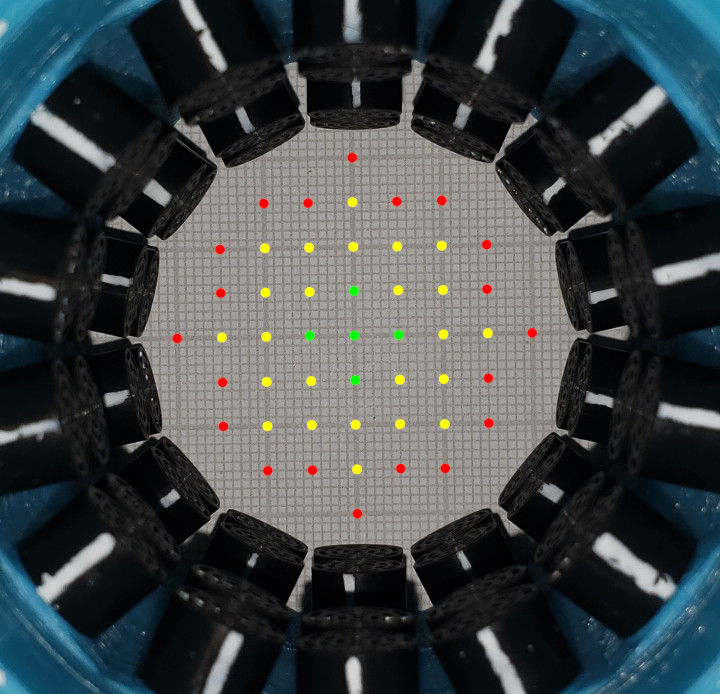}
    \caption{This chart represents the acoustic manipulator's \textit{basin of attraction}, the area that objects can be levitated or translated from to be levitated. The acoustic manipulator has been tested to successfully perform the picking action on objects at test point positions represented by the green dots and pull objects from the yellow dots to the center to be levitated. The red dots represent positions that were tested where objects could not be pulling into a position capable of picking. Each grid unit is 1mm and each dot is spaced 5mm apart.}
    \label{BasinOfAttraction}
\end{figure}

\section{Results}
\label{Sec:Results}

The acoustic picking device is capable of picking up objects with a maximum dimension of 2mm or less. We tested our picking device using polystyrene balls ranging from 2mm to 1mm in diameter. The polystyrene balls are an example of an object which is small enough to be occluded by the parallel jaw gripper of the PR2 robot and soft enough that the PR2 will crush the balls in the gripper when trying to pick them up. The contact-less ultrasonic levitation device can also levitate objects such as plants, small insects, integrated circuit chips, and potato chip crumbs, see Fig. \ref{ItemsLevitated}. The mass of each object is given in Table. \ref{objectMassTable}.

\subsection{Contact-less Ultrasonic General Purpose Picking Device}

The acoustic picking device serves as a modular manipulator for the PR2 robot. By grasping the acoustic manipulation device, the PR2 is able to manipulate objects much smaller than it could with its gripper. Additionally, the parallel jaw gripper of the PR2 robot securely fits into a groove built into the handle of the levitation device, allowing the PR2 gripper to mate with the acoustic manipulation device to optimize ease of use. This easy to grasp design allows the PR2 to pick up the device to manipulate small objects and set down the manipulator to regain the gripper's original functionality. This system functions as a practical extension to the manipulation capabilities of general purpose robotic grippers.

The picking procedure for the double ring levitation device takes advantage of both phase manipulation and selective activation and deactivation. Fig. \ref{MultiFramePicking} shows the sequence of ring activity and phase shifts which allow the object to be translated from the surface of the table, to the center of the top pair of rings. To test the picking action of the device, the gripper was first fitted with the ultrasonic manipulation device. The picking device was then placed on the table around a polystyrene ball. The device is turned on and the ring state and phase sequence in Fig. \ref{MultiFramePicking} is performed manually by a human operator, allowing the ball to be picked up. Once trapped in the upper ring, the acoustic levitation device can be picked up and moved.

In an ideal scenario, the robot would be capable of positioning the acoustic manipulation device with the object directly in the center of the cylindrical transducer array, however, robots designed for larger manipulation tasks, such as the PR2 used for testing, have a positioning error larger than the size of the target object. The acoustic manipulation device compensates for positioning error by increasing the ``grasping" range of the robot. For contact-less manipulation via acoustic levitation, we call this the \textit{basin of attraction}, shown in Fig. \ref{BasinOfAttraction}. This basin refers to the area in which objects inside the cylinder can be either picked up or pulled by the acoustic field toward the inner area where the picking action via acoustic levitation is possible. The extent of the basin of attraction is shown in green and yellow. The green area denotes positions that the acoustic levitation device can directly lift without lateral manipulation and the yellow denotes the area where the manipulator can laterally move objects into the green area from. To compensate for the 10mm positioning uncertainty of the PR2, the basin of attraction is about 30mm in diameter, an area of 706.86mm$^2$, allowing for objects to be picked up despite positioning misalignment.


\subsection{Acoustic Manipulation Device for Object Sorting}


When mounted to the PR2 robot, we tested the general usability of the acoustic manipulator with an object sorting task. In the test, a polystyrene ball of either red or blue was placed on the table in front of the robot. The PR2 first picked up the ball using the acoustic manipulator as described above. The PR2 then brought the picker up to a camera and extracted visual features within the picker. The PR2 successfully localized the object and identified its' color, then moved the picker above the corresponding color circle and dropped the polystyrene ball into the correct area. This experiment suggested that (1) the weight and size of this device does not impact the robot arm's normal functionality, (2) the contact-less grasp was strong enough to resist air turbulence generated by robot movements, (3) the contact-less grasp provides unoccluded object view for visual inspection.

\subsection{Limitations of the Acoustic Manipulator}

Although the contact-less ultrasonic manipulator can perform the picking action on polystyrene balls, the manipulation device has limitations and can be further improved to increase robustness and predictability, allowing the process to be automated and fully controlled by the robot. 

Additional elements, such as a feedback loop capable of localizing the object in both height and lateral position within the trap, would improve fine tuned and coordinated object movement. Movement planning and ultrasonic transducer phase calculation becomes easier to automate and more accurate with a localization method, enabling fine tuned rotation and translation of objects within the trap. 

Another improvement to the system also includes object geometry recognition and mapping to help optimize acoustic field strength based on target object size and shape. While picking objects are limited to small objects approximately 2mm in size, acoustic lift force could be further maximized to accommodate larger objects.

Designed for a controlled environment, variations in environmental variables can effect the performance of the device. One example of this is heat generated by the transducers. Variations in air temperature can change the air density within the trap, causing errors in device control.

\section{Conclusion}
\label{Sec:Conclusion}

This work presents an ultrasonic, contact-less manipulator designed to add millimeter scale object control to general purpose robots, like the PR2. The picking function is enabled by the acoustic field modeling technique, improving the simulated force dynamics inside the trap. With these simulations we were able to improve the accuracy of the system and inform the geometry of the device design. Likewise, the basin of attraction helped to improve the picking function by expanding the grasping range. Acoustic levitation and phase based object manipulation lends itself to robotics by compensating for the robot's positioning uncertainty and grasping force control. To make these actions more robust and fully automatic, additional improvements to the software and hardware could help with environmental factor immunity and increase object control. Implementing an object localization and feedback loop would help to improve this device's design and long term usage in performing more complicated tasks.

\section*{Acknowledgments} 
This work was funded in part by the Milton and Delia Zeutschel Professorship, and in part by National Science Foundation awards CNS-1823148 and EFMA-1832795.
\bibliographystyle{IEEEtran}
\bibliography{References}

\end{document}